\documentclass[11pt]{article}

\usepackage[margin=1in]{geometry}

\usepackage{xcolor}
\definecolor{tauorange}{HTML}{E07B3A}    
\definecolor{taudark}{HTML}{1F1812}      
\definecolor{taumuted}{HTML}{6B6055}     

\usepackage[colorlinks]{hyperref}
\hypersetup{
    colorlinks,
    linkcolor=tauorange,
    citecolor=tauorange,
    urlcolor=tauorange,
    filecolor=tauorange,
}

\usepackage{cite}
\usepackage{amsmath,amssymb,amsfonts,mathrsfs}
\usepackage{algorithmic}
\usepackage{graphicx}
\graphicspath{{images/}}
\usepackage{textcomp}
\usepackage{tcolorbox}
\tcbuselibrary{breakable}
\usepackage{bm}
\usepackage{booktabs}
\usepackage{verbatim}
\usepackage{url}
\usepackage{array}

\usepackage{tablefootnote}
\usepackage{threeparttable}
\usepackage{caption, subcaption}
\usepackage{titlesec}
\titleformat{\section}
  {\Large\bfseries\color{taudark}}
  {\color{tauorange}\thesection}{0.7em}{}
\titleformat{\subsection}
  {\large\bfseries\color{taudark}}
  {\color{tauorange}\thesubsection}{0.7em}{}
\titleformat{\subsubsection}
  {\normalsize\bfseries\color{taudark}}
  {\color{tauorange}\thesubsubsection}{0.7em}{}

\usepackage{fancyhdr}
\renewcommand{\headrulewidth}{0.4pt}

\renewcommand{\headrule}{\hbox to\headwidth{%
  \color{tauorange}\leaders\hrule height \headrulewidth\hfill}}
\usepackage{tikz}
\usepackage{tkz-euclide}

\usetikzlibrary{positioning}
\usetikzlibrary{shapes}
\usetikzlibrary{shapes.misc}
\usetikzlibrary{shapes.geometric}
\usetikzlibrary{plotmarks}
\usetikzlibrary{intersections}
\usetikzlibrary{calc}
\usetikzlibrary{fit}
\usetikzlibrary{patterns,tikzmark}
\usetikzlibrary{matrix,decorations.pathreplacing,calc}

\tikzset{cross/.style={cross out, draw, 
         minimum size=2*(#1-\pgflinewidth), 
         inner sep=0pt, outer sep=0pt}}

\newcommand{\vertiii}[1]{{\left\vert\kern-0.25ex\left\vert\kern-0.25ex\left\vert #1 
    \right\vert\kern-0.25ex\right\vert\kern-0.25ex\right\vert}}

\makeatletter
\renewcommand{\fps@figure}{htp}
\renewcommand{\fps@table}{htp}
\makeatother

\newcommand{\linkpill}[2]{%
  \href{#1}{%
    \fcolorbox{tauorange}{tauorange!8}{%
      \strut\;\textbf{#2}\;%
    }%
  }%
}

\def\BibTeX{{\rm B\kern-.05em{\sc i\kern-.025em b}\kern-.08em
    T\kern-.1667em\lower.7ex\hbox{E}\kern-.125emX}}

\begin{document}
\thispagestyle{plain}

\noindent\hfill

\vspace{0.6em}
{\color{tauorange}\rule{\textwidth}{1.2pt}}\\[-0.8em]
\begin{center}
  {\Huge\bfseries\color{taudark} MuJoCo-Drones-Gym\par}
  \vspace{0.25em}
  {\Large\color{taudark} A GPU-Accelerated Multi-Drone Simulator\\
  for Control and Reinforcement Learning\par}
\end{center}
{\color{tauorange}\rule{\textwidth}{1.2pt}}

\vspace{0.7em}
\begin{center}
  {\large\textbf{Manan Tayal}}\\[0.25em]
  {\small\color{taumuted} \href{https://tau-intelligence.com/}{TAU-Intelligence}\,\,$\bullet$\,\,%
   \href{mailto:robotics.tayalmanan@gmail.com}{robotics.tayalmanan@gmail.com}}\\[0.7em]
  \linkpill{https://github.com/tau-intelligence/MuJoCo-drones-gym}{Code}%
  \hspace{0.5em}%
  \linkpill{https://tau-intelligence.com/MuJoCo-Drones-Gym}{Web}
\end{center}

\vspace{0.9em}
\begin{tcolorbox}[
  breakable,
  colback=tauorange!4,
  colframe=tauorange,
  boxrule=0.6pt, arc=2mm,
  left=1.2em, right=1.2em, top=0.8em, bottom=0.8em,
]
\noindent{\bfseries\color{taudark}\large Abstract.}\quad
Robotic simulators are a cornerstone of modern research in aerial robotics,
serving both as a vehicle for the development of new control algorithms and
as the data source for training reinforcement learning (RL) policies. Yet,
existing quadcopter learning environments often face a trade-off between
physical fidelity, multi-agent support, and the throughput required by modern
deep RL pipelines.
In this paper, we present \textbf{MuJoCo-Drones-Gym}, an open-source
\texttt{Gymnasium}-compatible multi-drone environment built on top of the
\texttt{MuJoCo} physics engine. MuJoCo-Drones-Gym supports an arbitrary number of
Bitcraze Crazyflie 2.x nano-quadcopters and exposes a modular API for
selecting (i)~the physics model (rigid-body MuJoCo, explicit Python dynamics,
or any subset of ground effect, blade drag, and inter-drone downwash),
(ii)~the action interface (per-motor RPMs, collective normalized thrust,
velocity setpoints, or PID waypoint commands), and (iii)~the observation
space (kinematic state vectors, RGB / depth / segmentation cameras, or
neighbourhood adjacency information). A PettingZoo \texttt{ParallelEnv}
wrapper enables drop-in multi-agent reinforcement learning, while a suite of
seven task environments---hover, velocity tracking, multi-drone hover,
waypoint navigation, formation flight, gate racing, and a generic
multi-agent template---demonstrates the breadth of the interface. We
describe the environment design, the underlying physics and quadcopter
dynamics, and illustrate its use through control and learning examples
that mirror those of the closely related \emph{gym-pybullet-drones} project,
while taking advantage of MuJoCo's improved contact handling, rendering, and
parallelizability.
\end{tcolorbox}

\vspace{0.5em}

\section{Introduction}
\label{sec:introduction}

Open-source quadrotor simulators have become a \emph{de facto} benchmark
for reinforcement-learning (RL) and control research in aerial
robotics. Among them,
\emph{gym-pybullet-drones}~\cite{pybullet-drones} is the most widely
adopted: a Bullet-based Gym environment exposing the Bitcraze
Crazyflie~2.x and offering single- and multi-agent training
scripts. It demonstrated that a Python-first, pip-installable quadrotor
Gym---when coupled with well-tuned PID controllers and explicit
aerodynamic models---can serve as both a control-theory test-bed and an
RL benchmark, and the package has since been adopted as the comparison
baseline for a large body of follow-up work.

The underlying PyBullet engine, however, is now both slower and less
accurate than \emph{MuJoCo}~\cite{todorov2012mujoco} on the same
hardware, lacks first-class GPU vectorization, and is only partially
aligned with the modern \emph{Gymnasium}~\cite{gymnasium2024} and
\emph{PettingZoo}~\cite{pettingzoo2021} APIs that have replaced OpenAI
Gym as the community standard. Meanwhile, MuJoCo has become free and
open-source, ships with a built-in offscreen renderer
(\texttt{mujoco.Renderer}), and now exposes an XLA-compiled
GPU back end---\emph{MJX}~\cite{mjx2023}---that enables batched
simulation of thousands of environments on a single accelerator via
JAX~\cite{jax2018}.

\begin{figure}[t]
  \centering
  \includegraphics[width=0.85\columnwidth]{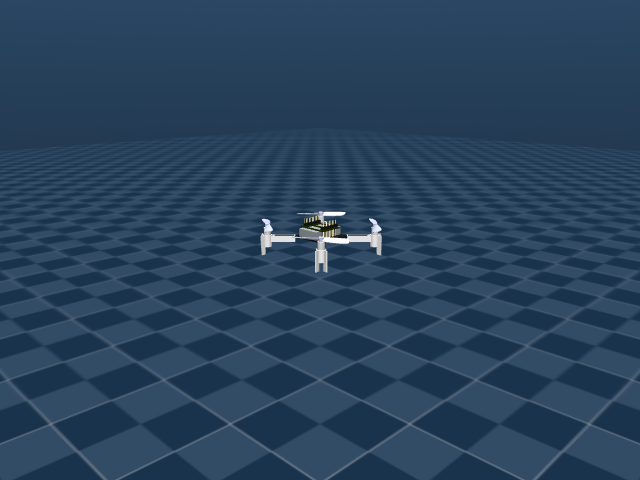}
  \caption{PID-controlled Bitcraze Crazyflie~2.x hovering at
  $z=1$~m in MJ-drones-gym, rendered with the built-in
  \texttt{track} camera mode
  (\texttt{BaseAviary.render(camera\_mode="track")}). Reproduced from
  \texttt{examples/pid.py::pid\_hover}.}
  \label{fig:hover-track}
\end{figure}

This paper introduces \textbf{MuJoCo-drones-gym}~\cite{mj_drones_gym}, an
open-source multi-quadcopter learning environment that ports the
gym-pybullet-drones design philosophy to MuJoCo and extends it.
Figure~\ref{fig:hover-track} shows a representative hover snapshot.
The key contributions of the package are:

\begin{enumerate}
  \item \textbf{A MuJoCo back end} with six selectable physics modes
        (pure MuJoCo, ground effect, drag, downwash, all-combined,
        plus an explicit-integration fallback) and
        \texttt{xfrc\_applied}-based rotor forcing
        (Section~\ref{subsec:method-dyn}).
  \item \textbf{Seven task environments} out of the box
        (\texttt{HoverAviary}, \texttt{VelocityAviary},
        \texttt{MultiHoverAviary}, \texttt{FlyThroughAviary},
        \texttt{FormationAviary}, \texttt{RaceAviary}, and
        \texttt{MultiAgentAviary}), summarized in
        Section~\ref{sec:experiments}.
  \item \textbf{A PettingZoo \texttt{ParallelEnv} wrapper} for native
        multi-agent RL on top of the same physics engine.
  \item \textbf{A cascaded PID stack}
        (\texttt{PIDControl}\,/\,\texttt{DSLPIDControl}) tuned for the
        MuJoCo Crazyflie dynamics, used both as a baseline controller
        and as the inner loop of the \texttt{PID}, \texttt{VEL}, and
        \texttt{ATTITUDE} action types.
  \item \textbf{Composable wrappers}: a Dryden-turbulence-plus-gust
        wind field, procedural obstacles
        (\texttt{FOREST}\,/\,\texttt{URBAN}\,/\,\texttt{INDOOR}\,/\,%
        \texttt{RANDOM}\,/\,\texttt{GATES}\,/\,\texttt{CUSTOM}), and an
        automatic curriculum that scales difficulty from per-episode
        metrics.
  \item \textbf{Three drone models}---Bitcraze \texttt{CF2X} and
        \texttt{CF2P} (X- and $+$-configuration Crazyflie~2.x) and a
        250\,g \texttt{RACE} class---with system-identified parameters
        from~\cite{forster2015system}.
  \item \textbf{GPU vectorization} via
        \texttt{multi\_drone\_mujoco.vectorized}: an MJX-based
        \texttt{MJXVectorAviary} that runs thousands of parallel
        environments in a single \texttt{jax.vmap}\,/\,\texttt{jit}
        call, plus an \\\texttt{MJXVecEnvGymWrapper} that exposes a
        Gymnasium \texttt{VectorEnv} for Stable-Baselines3 and other
        numpy-based RL libraries (Section~\ref{subsec:method-mjx}).
  \item \textbf{Vendored Crazyflie~2.x meshes} under
        \texttt{multi\_drone\_mujoco/assets/cf2/}, so installation
        does not require cloning the full
        \emph{MuJoCo Menagerie}~\cite{menagerie2022}; the mesh
        directory is overridable via the
        \texttt{MJ\_DRONES\_CF2\_ASSETS} environment variable.
\end{enumerate}

The rest of the paper is organized as follows.
Section~\ref{sec:background} positions MuJoCo-drones-gym in the
quadrotor-simulator landscape.
Section~\ref{sec:methodology} details the package internals---dynamics,
aerodynamics, physics modes, software architecture, observation and
action spaces, optional wrappers, and the MJX GPU back end.
Section~\ref{sec:experiments} walks through the task suite, the PID and
geometric controllers, and representative RL and disturbance-rejection
use cases.
Section~\ref{sec:conclusions} provides a head-to-head comparison with
gym-pybullet-drones and concludes.

\section{Background and Related Work}
\label{sec:background}

The space of quadrotor simulators spans high-fidelity tools (Gazebo,
AirSim, Flightmare) and lightweight RL environments
(\emph{gym-pybullet-drones}, the original \texttt{gym} MuJoCo suite).
MuJoCo-drones-gym sits in the latter category but specifically targets the
gap left by \emph{gym-pybullet-drones}~\cite{pybullet-drones}, on a
more modern physics back end with native support for GPU
vectorization.

\paragraph{gym-pybullet-drones~\cite{pybullet-drones}}
The first multi-quadrotor Gym environment with realistic aerodynamics:
the Shi~\emph{et~al.}~\cite{shi2019neural} ground-effect model, the
Forster~\cite{forster2015system} blade drag, and Zhou's downwash model
from the Dynamic Systems Lab (DSL). Its API, aerodynamic models, and
PID architecture are the direct ancestors of those used here, so that
controllers and policies port across the two environments with minimal
effort.

\paragraph{safe-control-gym~\cite{yuan2022safecontrolgym} and
CrazySwarm2}
The former targets classical and safe learning-based control with a
small Crazyflie environment built on PyBullet; the latter focuses on
ROS\,2 firmware-in-the-loop and real-hardware integration. Both have
comparatively simpler physics than MuJoCo-drones-gym and do not provide a
first-class multi-agent or GPU-vectorized interface.

\paragraph{Flightmare~\cite{song2020flightmare}}
Couples Unity-based rendering with a fast custom dynamics engine and
exposes a Gym-style RL interface for single-agent control. It is
heavier to install than a pip-only package and lacks a native
multi-agent API.

\paragraph{OmniDrones~\cite{xu2024omnidrones}}
A recent Isaac Sim / Isaac Lab based environment that targets
large-scale GPU RL on multi-quadrotor tasks. It delivers excellent
throughput but is tightly coupled to the NVIDIA Isaac stack and to
proprietary GPU drivers, which restricts deployment to a narrower set
of platforms than a pure JAX/MJX pipeline.

MuJoCo-drones-gym aims to occupy the same niche as gym-pybullet-drones
---lightweight installation, pure Python, Gymnasium-standard---while
adopting a more modern physics back end (MuJoCo / MJX), the
PettingZoo \texttt{ParallelEnv} contract for MARL, and a richer set of
task templates, controllers, and disturbance wrappers. The detailed
feature comparison is deferred to Table~\ref{tab:mj-vs-pyb} in
Section~\ref{sec:conclusions}.

\section{Methodology}
\label{sec:methodology}

This section describes the inner workings of MuJoCo-drones-gym.
Section~\ref{subsec:method-dyn} introduces the quadrotor dynamics and
aerodynamic models;
Section~\ref{subsec:method-modes} catalogs the six physics modes;
Section~\ref{subsec:method-models} the three drone models;
Section~\ref{subsec:method-arch} the software architecture, public API,
and observation/action spaces;
Section~\ref{subsec:method-wrappers} the optional wrappers (wind,
obstacles, curriculum, domain randomization); and finally
Section~\ref{subsec:method-mjx} the GPU-vectorized MJX back end.

\subsection{Quadrotor Dynamics}
\label{subsec:method-dyn}

\subsubsection{Notation}
Each drone has state
$\mathbf{x} = [\mathbf{p}, \mathbf{q}, \mathbf{v}, \bm{\omega}]^\top
\in \mathbb{R}^{13}$, where $\mathbf{p} \in \mathbb{R}^3$ is the
world-frame position, $\mathbf{q} \in \mathbb{R}^4$ the body-to-world
quaternion (in MuJoCo's $[w, x, y, z]$ convention), $\mathbf{v}$ the
linear velocity and $\bm{\omega}$ the angular velocity. The control
input is the vector of motor speeds
$\mathbf{n} = [n_0, n_1, n_2, n_3]^\top$ in RPM (radians per second
after the appropriate conversion).

\subsubsection{Motor model}
Per-motor thrust and reaction torque are quadratic in motor speed,
\begin{equation}
F_i = k_f\, n_i^2,\qquad
\tau_{z,i} = (-1)^{m_i}\, k_m\, n_i^2,
\label{eq:motor}
\end{equation}
with the sign $m_i \in \{0, 1\}$ alternating to produce yaw control.
For the X-configuration (\texttt{DroneModel.CF2X} and
\texttt{DroneModel.RACE}) the body-frame roll and pitch torques are
\begin{align}
\tau_x &= \tfrac{L}{\sqrt{2}}\,(F_0 + F_1 - F_2 - F_3),\\
\tau_y &= \tfrac{L}{\sqrt{2}}\,(-F_0 + F_1 + F_2 - F_3),
\end{align}
and the collective thrust is $T = \sum_i F_i$. The $+$-configuration
(\texttt{DroneModel.CF2P}) replaces the $L/\sqrt 2$ factor with $L$ on
a different pair of motors.

\subsubsection{Aerodynamic effects}
Three optional effects can be toggled independently via the
\texttt{Physics} enum of Section~\ref{subsec:method-modes}.

\paragraph{Ground effect~\cite{shi2019neural}}
At low altitude each rotor sees an additional vertical thrust
\begin{equation}
\Delta F_z^{\text{gnd}} \;=\; C_{\text{gnd}} \sum_i (k_f n_i^2)
\left(\frac{r_p}{4\,h}\right)^{\!2},
\label{eq:ground}
\end{equation}
where $h = \max(p_z, h_{\text{clip}})$ and $h_{\text{clip}}$ is derived
from the propeller radius and the model's maximum thrust to avoid the
singularity as $h \to 0$. The effect is gated on $|\phi|,|\theta| < \pi/2$
so an inverted drone does not gain spurious thrust
(\texttt{\_groundEffect}).

\paragraph{Blade drag~\cite{forster2015system}}
A body-frame, velocity-proportional drag is scaled by the sum of motor
speeds,
\begin{equation}
\mathbf{F}_{\text{drag}} \;=\; -\,\mathrm{diag}(c_{xy}, c_{xy}, c_z)
\left(\sum_i \tfrac{2\pi n_i}{60}\right)\,\mathbf{v}_{\text{body}},
\label{eq:drag}
\end{equation}
and rotated back to the world frame via the body's $3\times 3$
rotation matrix (\texttt{\_drag}).

\begin{figure}[t]
  \centering
  \includegraphics[width=0.85\columnwidth]{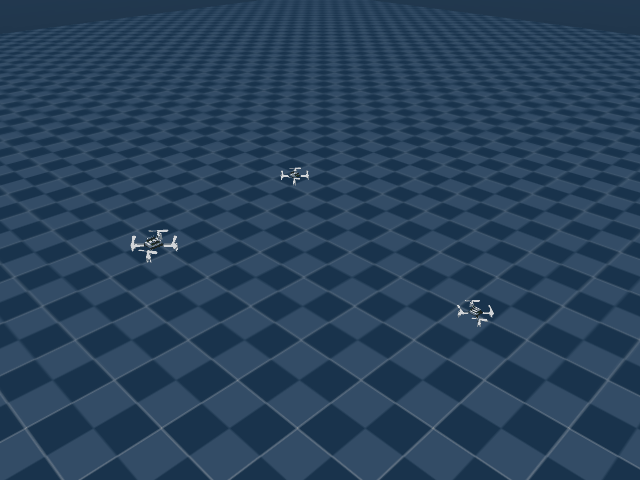}
  \caption{Two stacked Crazyflies under
  \texttt{Physics.MJC\_GND\_DRAG\_DW}: the bottom drone increases its
  thrust to compensate for the wake of the top drone
  (\texttt{examples/downwash.py}).}
  \label{fig:downwash}
\end{figure}

\paragraph{Downwash}
For every ordered pair $(i, j)$ with drone $i$ above $j$ and lateral
separation $\Delta_{xy} < 10$~m, the lower drone receives a downward
force
\begin{equation}
\Delta F_z^{\text{dw}} \;=\; -\alpha\,
\exp\!\left(-\tfrac{1}{2}(\Delta_{xy}/\beta)^2\right),
\label{eq:downwash}
\end{equation}
with $\alpha = C_1 (r_p/(4\,\Delta z))^2$,
$\beta = C_2\,\Delta z + C_3$, and the Crazyflie~2.x coefficients
$(C_1, C_2, C_3) = (2267.18,\ 0.16,\ -0.11)$
(\texttt{\_downwash}). Figure~\ref{fig:downwash} shows the resulting
transient on two vertically stacked drones.

\subsection{Physics Modes}
\label{subsec:method-modes}

The six selectable modes are listed in Table~\ref{tab:physics}. The
\texttt{MJC*} family applies rotor forces and torques through MuJoCo's
external-force interface
($\texttt{data.xfrc\_applied}[\text{body}, :3] = \mathbf{F}_{\text{world}}$,
\\$\texttt{data.xfrc\_applied}[\text{body}, 3:] = \bm{\tau}_{\text{world}}$)
and lets \texttt{mj\_step} integrate the rigid-body dynamics. The
\texttt{DYN} mode instead integrates the body explicitly in Python
(Euler), writes the resulting pose into \texttt{qpos}\,/\,\texttt{qvel}
before calling \texttt{mj\_forward}, and mirrors the explicit
integrator of \emph{gym-pybullet-drones}; it is useful for validating
the MuJoCo integrator against an analytical reference. Quaternions are
propagated in \texttt{\_integrateQ} via the closed-form exponential
map
\begin{equation}
\begin{aligned}
q(t{+}\Delta t) = \big[&\cos\!\big(\tfrac{\|\omega\|\Delta t}{2}\big)\,I \\
 &+ \tfrac{2}{\|\omega\|}\sin\!\big(\tfrac{\|\omega\|\Delta t}{2}\big)\,
   \Lambda(\omega)\big]\,q(t),
\end{aligned}
\end{equation}
where $\Lambda(\omega)$ is the standard $4\times 4$ matrix of the
quaternion product with $\omega$.

\begin{table}[t]
\centering
\caption{Physics modes supported by \texttt{BaseAviary}.}
\label{tab:physics}
\renewcommand{\arraystretch}{1.15}
\begin{tabular}{lcccc}
\hline
\textbf{Mode}                 & \textbf{Integrator} & \textbf{Ground}
 & \textbf{Drag} & \textbf{Downwash}\\
\hline
\texttt{MJC}                  & MuJoCo RK4   & --   & --   & --   \\
\texttt{MJC\_GND}             & MuJoCo RK4   & \checkmark & -- & -- \\
\texttt{MJC\_DRAG}            & MuJoCo RK4   & -- & \checkmark & -- \\
\texttt{MJC\_DW}              & MuJoCo RK4   & -- & -- & \checkmark \\
\texttt{MJC\_GND\_DRAG\_DW}   & MuJoCo RK4   & \checkmark & \checkmark & \checkmark \\
\texttt{DYN}                  & Explicit Euler & -- & -- & -- \\
\hline
\end{tabular}
\end{table}

\subsection{Drone Models}
\label{subsec:method-models}

Three rigid-body models are bundled, all using the parameters
of~\cite{forster2015system} for the Crazyflies and a published
racing-class set for \texttt{RACE} (Table~\ref{tab:drone-models}). The
derived quantities $\texttt{HOVER\_RPM} = \sqrt{m g / (4 k_f)}$ (hover
rotor speed) and \texttt{MAX\_RPM} (motor saturation) are computed
in \texttt{BaseAviary.\_\_init\_\_} from $(m, g, k_f, T/W)$.

\begin{table}[t]
\centering
\caption{Drone model parameters
(\texttt{multi\_drone\_mujoco.utils.enums.DroneModel}).}
\label{tab:drone-models}
\renewcommand{\arraystretch}{1.15}
\begin{tabular}{lcc}
\hline
\textbf{Parameter} & \texttt{CF2X}\,/\,\texttt{CF2P} & \texttt{RACE} \\
\hline
mass $m$                  & 27\,g       & 250\,g      \\
arm length $L$            & 39.7\,mm    & 125\,mm     \\
thrust\,/\,weight         & 2.25        & 4.0         \\
$J_{xx}, J_{yy}$ (kg\,m$^2$)
                          & $1.4{\cdot}10^{-5}$  & $4.86{\cdot}10^{-4}$ \\
$J_{zz}$ (kg\,m$^2$)      & $2.17{\cdot}10^{-5}$ & $8.80{\cdot}10^{-4}$ \\
$k_f$ (N\,/\,(rad/s)$^2$) & $3.16{\cdot}10^{-10}$ & $1.28{\cdot}10^{-8}$ \\
$k_m$ (Nm\,/\,(rad/s)$^2$)& $7.94{\cdot}10^{-12}$ & $5.96{\cdot}10^{-10}$ \\
propeller radius $r_p$    & 23.25\,mm   & 63.5\,mm    \\
top speed (km/h)          & 30          & 100         \\
\hline
\end{tabular}
\end{table}

\subsection{Software Architecture and APIs}
\label{subsec:method-arch}

\subsubsection{Package layout}
The top-level Python package
\texttt{multi\_drone\_mujoco/} groups:

\begin{itemize}
  \item \texttt{envs/}: the seven task-specific Gymnasium environments
        (\texttt{base\_aviary.py},
        \texttt{hover\_aviary.py}, \texttt{velocity\_aviary.py},
        \texttt{multi\_hover\_aviary.py},
        \texttt{fly\_through\_aviary.py},
        \texttt{formation\_aviary.py},
        \texttt{race\_aviary.py}, and the PettingZoo wrapper
        \texttt{multi\_agent\_aviary.py}).
  \item \texttt{control/}: \texttt{PIDControl} (paper-grade gains) and
        \texttt{DSLPIDControl} (DSL-style PID with rate limiting and
        anti-windup).
  \item \texttt{wrappers/}: \texttt{wind.py}\,/\,\texttt{wind\_wrapper.py},
        \texttt{curriculum.py}, and \texttt{obstacles.py}.
  \item \texttt{utils/}: enums and a per-drone CSV+matplotlib logger.
  \item \texttt{vectorized/}: the MJX-based GPU back end
        (Section~\ref{subsec:method-mjx}).
  \item \texttt{assets/cf2/}: vendored Bitcraze CF2.x meshes (39 OBJ
        files + license), so the package installs without cloning the
        full MuJoCo Menagerie~\cite{menagerie2022}.
\end{itemize}

The package is \texttt{pip}-installable with dependency groups declared
in \texttt{pyproject.toml}: \texttt{[rl]} adds Stable-Baselines3,
\texttt{[marl]} adds PettingZoo, \texttt{[viz]} adds matplotlib and
Pillow, \texttt{[gpu]} adds \texttt{jax[cuda12]} and
\texttt{mujoco-mjx}, and \texttt{[all]} adds all of the above.

\subsubsection{Environment API}
All environments derive from \texttt{BaseAviary(gym.Env)} and follow
the standard Gymnasium 5-tuple step:
\begin{verbatim}
obs, reward, terminated, truncated, info = env.step(action)
\end{verbatim}
The orthogonal constructor options are
\texttt{drone\_model}, \texttt{num\_drones},
\texttt{physics}, \texttt{sim\_freq}, \texttt{ctrl\_freq},
\texttt{initial\_xyzs}\,/\,\texttt{initial\_rpys},
\texttt{obs\_type}, \texttt{act\_type}, \texttt{obstacles},
\texttt{vision\_attributes}, \texttt{gui}, \texttt{render\_mode}, and
\texttt{output\_folder}.
\texttt{sim\_freq} must be a positive integer multiple of
\texttt{ctrl\_freq}; the environment runs
$\texttt{sim\_freq}/\texttt{ctrl\_freq}$ MuJoCo sub-steps per
\texttt{env.step} call.

\subsubsection{Action types}
Selected via the \texttt{ActionType} enum and converted to rotor RPMs
in \texttt{\_preprocessAction}
(Table~\ref{tab:action-types}).

\begin{table}[t]
\centering
\caption{Action types exposed by \texttt{BaseAviary}.}
\label{tab:action-types}
\renewcommand{\arraystretch}{1.15}
\setlength{\tabcolsep}{4pt}
\footnotesize
\begin{tabular}{l c p{0.55\columnwidth}}
\hline
\textbf{Type} & \textbf{Shape} & \textbf{Description}\\
\hline
\texttt{RPM}        & $(4,)$ & raw motor speeds in $[0, \texttt{MAX\_RPM}]$ \\
\texttt{ONE\_D\_RPM}& $(1,)$ & normalized thrust $\in [-1, 1]$ to all 4 motors \\
\texttt{VEL}        & $(4,)$ & body-frame $[v_x, v_y, v_z, \omega_z]$, PID inner loop \\
\texttt{PID}        & $(4,)$ & absolute waypoint $[x, y, z, \psi]$, PID inner loop \\
\texttt{ATTITUDE}   & $(4,)$ & thrust $+$ roll $+$ pitch $+$ yaw-rate, mixer to RPM \\
\hline
\end{tabular}
\end{table}

\subsubsection{Observation types}
\texttt{KIN} returns a 20-dimensional per-drone state vector
$[\mathbf{p}, \mathbf{q}, \bm{\eta}, \mathbf{v}, \bm{\omega},
\mathbf{a}_{t-1}]$. \texttt{RGB} returns $(N, 48, 64, 4)$ uint8 frames
from each drone's onboard 60$^{\circ}$-FOV camera, positioned at
$[0.02, 0, 0]$ in the body frame. \texttt{KIN\_RGB} returns a
\texttt{gym.spaces.Dict} combining the two. Per-task observation
vectors are constructed from these primitives by each subclass's
\texttt{\_computeObs} (e.g., \texttt{HoverAviary} exposes only the
12-dim $[\mathbf{p}, \bm{\eta}, \mathbf{v}, \bm{\omega}]$ slice).

\subsection{Composable Wrappers}
\label{subsec:method-wrappers}

\subsubsection{Wind and turbulence}
The \texttt{WindField} object exposes
\texttt{get\_force(dt, position, velocity)}, returning a 3-D force in
newtons suitable for accumulation in \texttt{data.xfrc\_applied}. The
wind type is selected by \texttt{WindConfig.model}:

\begin{itemize}
  \item \texttt{CONSTANT}: steady wind with quadratic drag
        $\mathbf{F} = \tfrac{1}{2}\rho C_d A
        \|\mathbf{v}_{\text{rel}}\|^2 \hat{\mathbf{v}}_{\text{rel}}$.
  \item \texttt{GUST}: random impulses with a smooth $\sin(\pi s)$
        envelope and per-step trigger probability.
  \item \texttt{DRYDEN}: first-order colored noise driven by
        MIL-F-8785C length scales~\cite{mil8785c}, e.g.
        $L_u = h/(0.177 + 8.23{\cdot}10^{-4}\,h)^{1.2}$ at altitude
        $h$. The discrete-time recursion is
        $x_{t+\Delta t} = \alpha x_t + \sigma\sqrt{1 - \alpha^2}\,\xi$
        with $\alpha = e^{-\Delta t\,V/L}$.
  \item \texttt{SINUSOIDAL}: $\mathbf{F}(t) = A\,[\sin\omega t,
        \cos\omega t, 0.3\sin 2\omega t]$.
  \item \texttt{COMBINED}: sum of \texttt{CONSTANT + GUST + DRYDEN +
        SINUSOIDAL}.
\end{itemize}

\begin{figure}[t]
  \centering
  \includegraphics[width=0.95\columnwidth]{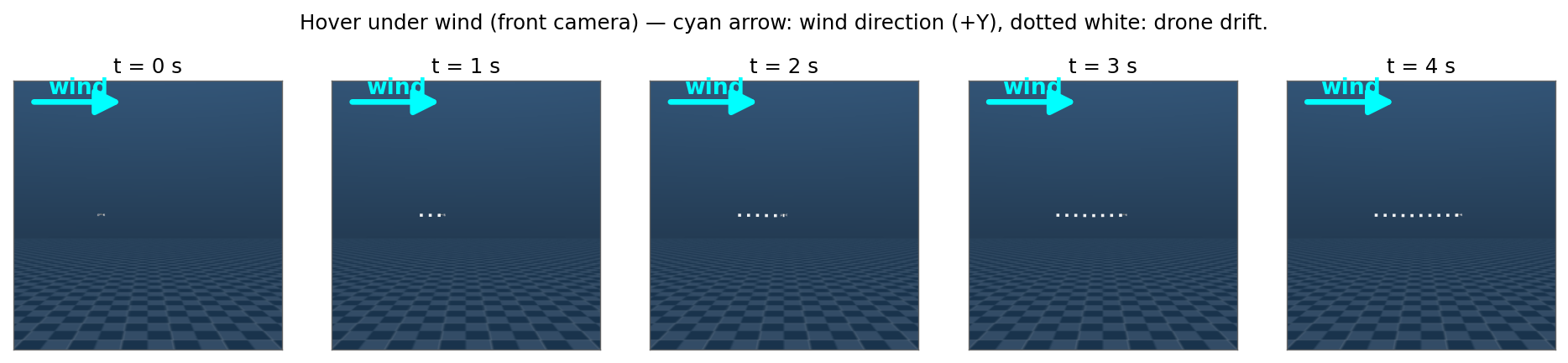}
  \caption{Hover at $z = 1$~m under \texttt{WindModel.COMBINED} (steady
  0.9~m/s along $+Y$ plus Dryden turbulence). The cyan arrow shows the
  wind direction; the dotted white trail is the drone's $YZ$ position.
  The PID controller of Section~\ref{subsec:exp-control} cannot fully
  reject the steady component without trimming and the drone drifts
  downwind at roughly $0.1$~m/s---exactly the kind of distribution that
  the curriculum (Section~\ref{subsubsec:method-curr}) and domain
  randomization (Section~\ref{subsubsec:method-dr}) wrappers expose to
  the policy during training.}
  \label{fig:wind}
\end{figure}

\texttt{WindWrapper(env, WindConfig(...))} composes with any
\texttt{BaseAviary} env, and applies the wind force on top of the
rotor wrench at every \texttt{env.step}. \texttt{BaseAviary} also
exposes \\\texttt{env.set\_wind(WindConfig(...))} for in-place attachment
without wrapping. Figure~\ref{fig:wind} illustrates a hover under
\texttt{COMBINED} wind.

\subsubsection{Procedural obstacles}
\begin{figure}[t]
  \centering
  \begin{minipage}{0.47\columnwidth}
    \includegraphics[width=\linewidth]{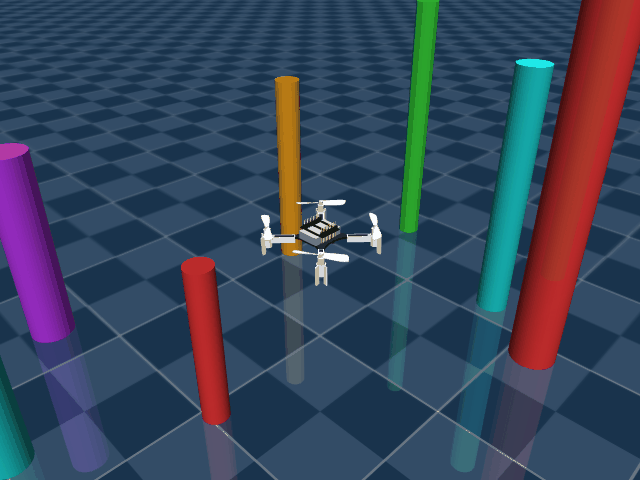}
  \end{minipage}\hfill
  \begin{minipage}{0.47\columnwidth}
    \includegraphics[width=\linewidth]{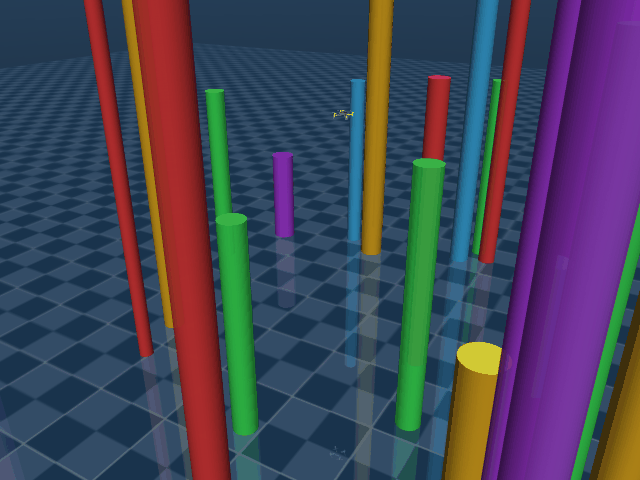}
  \end{minipage}
  \caption{\texttt{ObstacleType.FOREST} (left) and
  \texttt{ObstacleType.RANDOM} (right). The generated XML is
  concatenated into the world body of the aviary at instantiation
  time.}
  \label{fig:obstacles}
\end{figure}

\texttt{ObstacleConfig(obstacle\_type=ObstacleType.FOREST,
num\_obstacles=20, arena\_size=(3,3,2))} parametrizes one of the
procedural generators \texttt{FOREST}, \texttt{URBAN},
\texttt{INDOOR}, \texttt{RANDOM}, \texttt{GATES}, or \texttt{CUSTOM}.
The output is a list of
\texttt{Obstacle(geom\_type, position, size, rgba, euler)} records that
the wrapper splices into the MuJoCo XML before the model is loaded; a
\texttt{safe\_zone\_radius} keeps a sphere around each drone
collision-free at spawn time. Figure~\ref{fig:obstacles} shows two of
the procedural scenes.

\subsubsection{Curriculum learning}
\label{subsubsec:method-curr}
\texttt{CurriculumWrapper(env, difficulty\_fn)} maintains a discrete
$\text{level} \in [0, \text{num\_levels})$ and adjusts it after each
episode based on a moving average of \texttt{success\_rate},
\texttt{reward}, or \texttt{episode\_length}
(\texttt{CurriculumConfig.metric}). The user-supplied
\texttt{difficulty\_fn(env, level)} is called on every \texttt{reset}
and may mutate the wrapped env, e.g.\ raise the target altitude or
attach a wind field as the level grows.

\subsubsection{Domain randomization}
\label{subsubsec:method-dr}
\begin{figure}[t]
  \centering
  \includegraphics[width=\columnwidth]{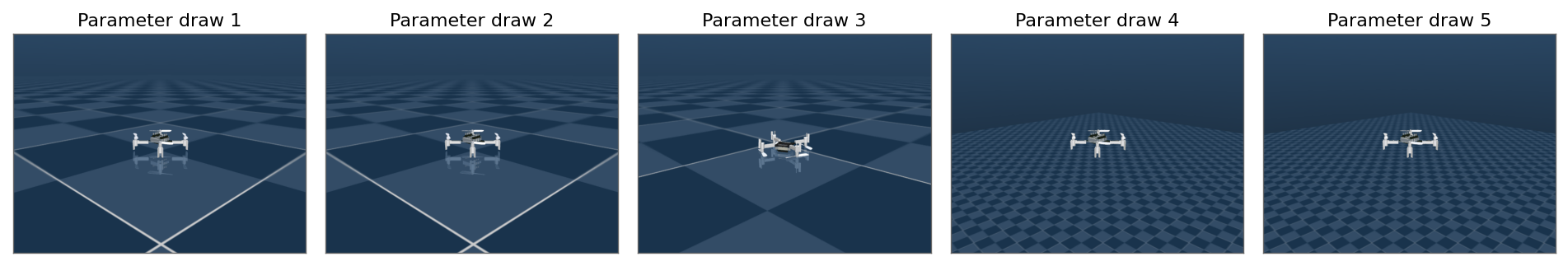}
  \caption{Five independent resets of the
  \texttt{DomainRandomizationWrapper}. Each panel is a different draw
  from the parameter distributions of Table~\ref{tab:dr-ranges}
  (different effective mass, drag, motor lag, sensor bias, $\dots$);
  within a single panel the parameters are constant, only the episode
  index changes across panels.}
  \label{fig:domain-randomization}
\end{figure}

A \texttt{DomainRandomizationWrapper} (exercised by
\texttt{tests/test\_features.py}) resamples a vector of physical and
sensor parameters at \emph{every} \texttt{reset()}, so that the policy
sees a wide distribution of dynamics during training while the
parameters stay fixed for the duration of each episode. This is the
standard sim-to-real recipe of~\cite{tobin2017domain,
openai2019solving} and the one we recommend for transferring a
MuJoCo-drones-gym policy to a real Crazyflie.
Table~\ref{tab:dr-ranges} summarizes the ranges we have found to work
well in practice for the \texttt{CF2X} model;
Figure~\ref{fig:domain-randomization} shows five independent draws.

\begin{table*}[!t]
\centering
\caption{Recommended domain-randomization ranges for \texttt{CF2X}.
Multiplicative factors are sampled per episode from
$\mathcal{U}(1{-}\alpha,\,1{+}\alpha)$; additive noise is per step
unless marked ``per ep.''.}
\label{tab:dr-ranges}
\renewcommand{\arraystretch}{1.15}
\footnotesize
\begin{tabular}{@{}l l l@{}}
\toprule
\textbf{Group} & \textbf{Parameter} & \textbf{Range}\\
\midrule
body     & mass $m$, inertia $J$                & $\times[0.8, 1.2]$ \\
         & arm length $L$                       & $\times[0.98, 1.02]$ \\
\midrule
aero     & thrust $k_f$, torque $k_m$           & $\times[0.85, 1.15]$ \\
         & drag $c_{xy}, c_z$, ground $C_{\text{gnd}}$, downwash $C_1$
                                                & $\times[0.7, 1.3]$ \\
\midrule
actuat.  & per-motor bias                       & $\times[0.95, 1.05]$ (4 draws) \\
         & motor lag $\tau_{\text{mot}}$        & $\mathcal{U}[5, 50]$~ms (per ep.) \\
         & action latency                       & $\mathcal{U}[0, 50]$~ms (per ep.) \\
         & \texttt{MAX\_RPM}                    & $\times[0.95, 1.05]$ \\
\midrule
sensors  & position / vel / RPY noise           & $\mathcal{N}(0,\,1\,\text{cm}\,/\,5\,\text{cm/s}\,/\,1^{\circ})$ \\
         & gyro noise                           & $\mathcal{N}(0,\,0.05\,\text{rad/s})$ \\
         & sensor latency                       & $\mathcal{U}[0, 20]$~ms (per ep.) \\
\midrule
env.     & gravity $g$                          & $\times[0.98, 1.02]$ \\
         & wind (\texttt{COMBINED})             & magnitude $\sim\mathcal{U}(0, 1)$~m/s \\
\midrule
init     & $\mathbf{p}_0$, RPY$_0$, $\mathbf{v}_0$
                                                & $\pm 0.2$~m / $\pm 0.1$~rad / $\pm 0.5$~m/s \\
\bottomrule
\end{tabular}
\end{table*}

In practice we recommend (i)~randomizing \emph{coefficients}
($m$, $k_f$, $k_m$, drag) rather than \emph{forces}, so the policy
learns the inverse mapping rather than memorising disturbances;
(ii)~always including motor lag and action latency, which are usually
the single biggest unmodelled effects on small quadrotors;
(iii)~coupling DR with a sticky per-episode bias on the observation,
so the policy is forced to filter rather than memorise; and
(iv)~treating wind as part of DR at training time via
\texttt{WindModel.COMBINED}, rather than as a separate wrapper.

\subsection{GPU-Vectorized Back End (MJX)}
\label{subsec:method-mjx}

Alongside the CPU \texttt{BaseAviary}, MuJoCo-drones-gym ships a fully
GPU-vectorized back end, \texttt{MJXVectorAviary}, that runs thousands
of independent drone simulations in parallel on a single GPU. It is
implemented on top of \emph{MuJoCo MJX}~\cite{mjx2023}---the
XLA-compiled re-implementation of MuJoCo---and the JAX~\cite{jax2018}
array library, so that the aviary can be batched with \texttt{jax.vmap}
for population-based search and for JAX-native pipelines such as
PureJaxRL.

At instantiation time a minimal MJCF (no visual meshes, to keep XLA
compilation tractable) is built and converted into a batched
\texttt{mjx.Model} via \texttt{mjx.put\_model}. Per-environment
simulation state lives in a JAX \texttt{NamedTuple} (\texttt{MJXState})
carrying the batched \texttt{mjx.Data}, a step counter, a PRNG key, and
a boolean \texttt{done}. A single-env \texttt{\_single\_step}
method---written in pure JAX---maps the normalized action to RPMs,
builds the per-rotor thrust and torques via~\eqref{eq:motor}, applies
them through \texttt{xfrc\_applied}, and advances the physics by
$f_{\text{sim}}/f_{\text{ctrl}}$ MJX sub-steps via
\texttt{jax.lax.scan(mjx.step, ...)}. It is then lifted to the batch
dimension with
\begin{verbatim}
self._step_fn  = jit(vmap(self._single_step))
self._reset_fn = jit(vmap(self._single_reset))
\end{verbatim}
so that a single XLA kernel fans out across all \texttt{num\_envs}
environments. The default configuration is \texttt{num\_envs}\,$=4096$,
but the only practical limit is GPU memory.

The public interface mirrors the Gymnasium contract while remaining
JAX-native:
{\footnotesize
\begin{verbatim}
from multi_drone_mujoco.vectorized \
    import MJXVectorAviary

env   = MJXVectorAviary(num_envs=4096, task="hover")
rng   = jax.random.PRNGKey(0)
state = env.reset(rng)
state, obs, reward, done, info = \
    env.step(state, action)    # action: (N,4)
\end{verbatim}
}
For compatibility with numpy-based RL libraries an
\texttt{MJXVecEnvGymWrapper} bridges the JAX-native interface to a
Gymnasium \texttt{VectorEnv}, transparently performing the
GPU\,$\to$\,CPU transfers on each \texttt{step()} call. When the
policy itself is written in JAX, the wrapper can be bypassed and the
entire environment--policy--optimizer loop kept on device, eliminating
host\,/\,device traffic and unlocking the bulk of the GPU's compute.
The MJX back end currently supports the \texttt{hover},
\texttt{stabilize}, and \texttt{track} tasks with the 12-dim per-drone
observation and the normalized 4-D RPM action; aerodynamic effects are
intentionally omitted in this back end to preserve XLA compilability
and constant-shape requirements, and the CPU \texttt{BaseAviary}
remains the reference implementation when those effects are required.

\section{Tasks, Control, and Examples}
\label{sec:experiments}

This section walks through (i)~the seven task environments shipped with
the package (Section~\ref{subsec:exp-tasks}), (ii)~the available control
stack---cascaded PID, DSL-style PID, and an SE(3) geometric controller
for aggressive trajectories
(Section~\ref{subsec:exp-control}), and (iii)~representative
reinforcement learning use-cases (Section~\ref{subsec:exp-examples}).

\subsection{Task Suite}
\label{subsec:exp-tasks}

All tasks default to \texttt{Physics.MJC} and
\texttt{ActionType.RPM} normalized to $[-1, 1]$ and mapped via
\texttt{BaseAviary.\_normalizedActionToRPM}.
Table~\ref{tab:task-summary} summarizes the per-task observation
and action dimensions, episode lengths, and default scenarios.

\begin{table}[t]
\centering
\caption{Default scenario parameters of the seven task environments.}
\label{tab:task-summary}
\renewcommand{\arraystretch}{1.15}
\begin{tabular}{lcccc}
\hline
\textbf{Env} & $N$ & \textbf{Episode (s)} & \textbf{Action} & \textbf{Obs}\\
\hline
\texttt{HoverAviary}     & 1 & 10 & 4 (norm.\ RPM) & 12 \\
\texttt{VelocityAviary}  & 1 & 10 & 4 (norm.\ RPM) & 16 \\
\texttt{MultiHoverAviary}& 2 & 10 & $4N$           & $13N$ \\
\texttt{FlyThroughAviary}& 1 & 20 & $4N$           & $18N$ \\
\texttt{FormationAviary} & 3 & 30 & $4N$           & $18N$ \\
\texttt{RaceAviary}      & 1 & 30 & $4N$           & $21N$ \\
\texttt{MultiAgentAviary}& 3 & 10 & 4 per agent    & 13 per agent \\
\hline
\end{tabular}
\end{table}

\subsubsection{\texttt{HoverAviary} --- single-drone hover}
The drone must hover at $z = \texttt{TARGET\_HEIGHT}$ (default
$1.0$~m), with $(x, y) = (0, 0)$ and a level body. The reward is
\begin{align}
r =\;&-|z - z^*| - 0.1\,\|(x, y)\| - 0.05\,\|\mathbf{v}\|
      - 0.05\,(|\phi| + |\theta|) \notag\\
      &+ 0.5\,\mathbb{1}_{\text{at-target}}
      - 100\,\mathbb{1}_{\text{crash}},
\end{align}
with termination when $z<0$, $\max(|\phi|, |\theta|) > \pi/2$, or
$z > 3$~m, and truncation at $\texttt{EPISODE\_LEN\_SEC}=10$.
Figure~\ref{fig:hover-track} shows a representative roll-out.

\subsubsection{\texttt{VelocityAviary} --- velocity tracking}
A 16-dim observation $[\mathbf{p}, \bm{\eta}, \mathbf{v}, \bm{\omega},
v_x^*, v_y^*, v_z^*, \omega_z^*]$, a 4-dim normalized-RPM action, and
reward $r = -\|\mathbf{v} - \mathbf{v}^*\| - 0.1\,|\omega_z -
\omega_z^*| - 0.05\,(|\phi| + |\theta|) + 0.5\,
\mathbb{1}_{\text{tracked}}$. The target is resampled per episode from
$\mathcal{U}(-0.5, 0.5)^3 \times \mathcal{U}(-0.25, 0.25)$~m/s.

\subsubsection{\texttt{MultiHoverAviary} --- multi-drone hover}
$N$ drones initialized on a horizontal line at $z = 0.1$~m must each
reach a distinct target altitude
(default \texttt{numpy.linspace(0.7, 1.2, N)}). Observation is $13N$
(kinematic state plus target altitude per drone), action is $4N$
normalized RPMs, and the reward is the summed per-drone hover reward of
\texttt{HoverAviary}, with the lateral-error term measured against each
drone's spawn position.

\subsubsection{\texttt{FlyThroughAviary} --- waypoint chasing}
A drone must visit a list of 3-D waypoints (default
$[(0,0,1), (1,0,1), (1,1,1.5), (0,1,1), (0,0,0.5)]$). The observation
extends \texttt{KIN} with the next waypoint and its position relative to
the body. Reward: $+10$ on waypoint capture,
$-0.1\,\|d_{\text{wp}}\|$ shaping, $-0.01\,\|\bm{\omega}\|$ smoothness.
Truncates at 20~s.

\subsubsection{\texttt{FormationAviary} --- formation flight}
\begin{figure}[t]
  \centering
  \includegraphics[width=0.85\columnwidth]{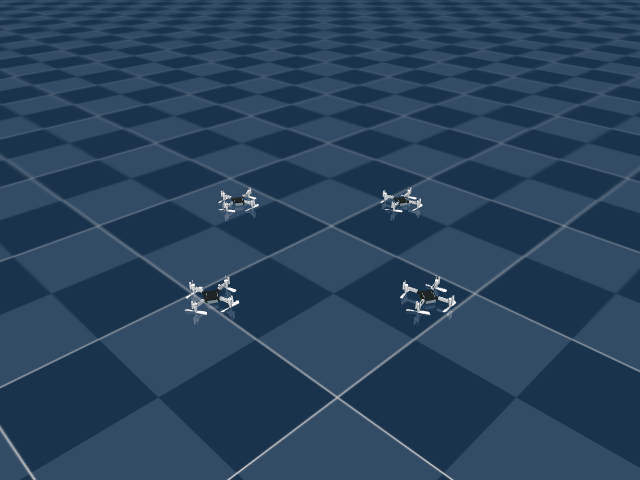}
  \caption{\texttt{MultiHoverAviary}\,/\,\texttt{FormationAviary}-style
  multi-drone scene.}
  \label{fig:formation}
\end{figure}

$N$ drones must maintain a fixed formation (default: equispaced points
on a circle of radius 0.3~m) while the formation centre tracks a closed
polygonal path. The reward sums the per-drone position tracking with an
inter-agent penalty
$-0.5 \sum_{i<j} |d_{ij}^{\text{actual}} - d_{ij}^{\text{desired}}|$.
Figure~\ref{fig:formation} shows a typical scene.

\subsubsection{\texttt{RaceAviary} --- multi-gate racing}
Six gates form a closed track that the drone must complete twice as
fast as possible. Reward: $+20$ per gate, $+2\,\|\mathbf{v}\|$ speed
bonus, $-0.05\,d_{\text{gate}}$ shaping. Episode caps at 30~s.

\subsubsection{\texttt{MultiAgentAviary} --- PettingZoo \texttt{ParallelEnv}}
A thin wrapper that turns \texttt{MultiHoverAviary} into a PettingZoo
parallel environment. Agents are addressed as
\texttt{drone0, drone1, \dots}; the per-agent observation is 13-D, the
per-agent action is 4-D RPMs, and the per-agent reward is the per-drone
hover reward of \texttt{MultiHoverAviary}. This is the entry point for
MARL training with libraries such as \texttt{MARLlib},
\texttt{RLlib}~\cite{liang2018rllib}, or \texttt{BenchMARL}.

{\footnotesize
\begin{verbatim}
from multi_drone_mujoco.envs.multi_agent_aviary \
    import MultiAgentAviary

env = MultiAgentAviary(num_drones=3)
obs, infos = env.reset()
actions = {agent: env.action_space(agent).sample()
           for agent in env.agents}
obs, rewards, terms, truncs, infos = env.step(actions)
\end{verbatim}
}

\begin{figure}[t]
  \centering
  \includegraphics[width=\columnwidth]{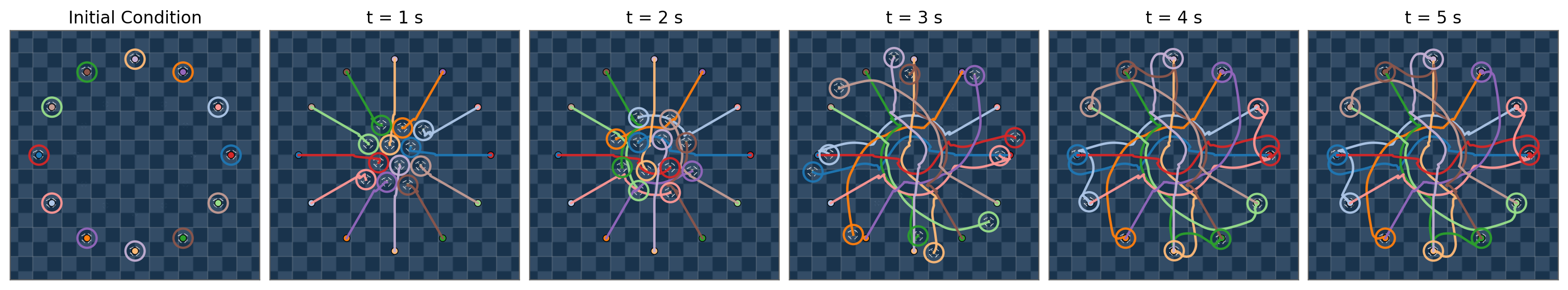}
  \caption{\texttt{MultiAgentAviary}-style benchmark: 12 \texttt{CF2X}
  drones with antipodal goals navigating in MuJoCo. Top-down snapshots
  at $t = 0, 1, 2, 3, 4, 5$~s; the colored rings are the safety radius
  and the small dots mark each drone's goal. The drones swirl through
  the central region while avoiding each other and converge on their
  antipodal goals.}
  \label{fig:marl}
\end{figure}

Figure~\ref{fig:marl} shows a 12-drone antipodal-navigation benchmark
built on top of \texttt{MultiAgentAviary}: each drone must reach the
antipodal point on a circle while avoiding the others. Such
collision-avoidance behaviour can either be shaped through reward terms
or enforced explicitly by wrapping the policy with a safety filter
such as a collision-cone control-barrier-function
constraint~\cite{tayal2024control}; both routes plug into the same
\texttt{ParallelEnv} contract.

\subsection{Control}
\label{subsec:exp-control}

\subsubsection{Cascaded PID --- \texttt{PIDControl}}
\texttt{PIDControl} implements the standard cascaded
position\,/\,attitude\,/\,motor-mixer architecture
of~\cite{luis2016design}. Its constructor reads the inertial constants
from a \texttt{BaseAviary} instance (or falls back to the published
\texttt{CF2X} values). The position loop yields the desired
acceleration
\begin{equation}
\mathbf{a}^*
 = K_p(\mathbf{p}^* - \mathbf{p})
 + K_i\!\int\!(\mathbf{p}^* - \mathbf{p})\,dt
 + K_d(\mathbf{v}^* - \mathbf{v})
 + g\,\hat{\mathbf{z}},
\end{equation}
with the default gains of Table~\ref{tab:pid-gains}.
The desired thrust $T = m\,\|\mathbf{a}^*\|$ and a small-angle attitude
target are then converted to roll\,/\,pitch\,/\,yaw torques by the
attitude PID, and the allocation matrix
\[
\begin{bmatrix} T\\ \tau_x \\ \tau_y \\ \tau_z \end{bmatrix} =
M\,
\begin{bmatrix} k_f n_0^2 \\ k_f n_1^2 \\ k_f n_2^2 \\ k_f n_3^2 \end{bmatrix}
\]
is inverted (with torque clamping to keep \texttt{numpy.linalg.solve}
away from motor saturation) to recover the per-motor RPMs. The
\texttt{computeControl} method returns
\texttt{(rpm, pos\_error, yaw\_error)}.

\begin{table*}[!t]
\centering
\caption{Default \texttt{PIDControl} gains for \texttt{CF2X} (tuned for
the MuJoCo dynamics in \texttt{pid\_control.py}).}
\label{tab:pid-gains}
\renewcommand{\arraystretch}{1.15}
\begin{tabular}{lcccc}
\hline
\textbf{Loop} & $K_p$ & $K_i$ & $K_d$ & \textbf{Integral clamp}\\
\hline
position  & $[0.4, 0.4, 1.0]$
          & $[0.01, 0.01, 0.01]$
          & $[0.9, 0.9, 2.0]$
          & $[-2, 2]$~m \\
attitude  & $[2{\cdot}10^{-3}, 2{\cdot}10^{-3}, 1{\cdot}10^{-3}]$
          & $[0, 0, 1{\cdot}10^{-4}]$
          & $[5{\cdot}10^{-4}, 5{\cdot}10^{-4}, 2{\cdot}10^{-4}]$
          & $[-0.5, 0.5]$ \\
\hline
\end{tabular}
\end{table*}

\subsubsection{\texttt{DSLPIDControl}}
\texttt{DSLPIDControl} retunes the position and attitude gains, halves
the integral clamps for anti-windup, and adds rate limits on the
position setpoint ($2.0$~m/s) and yaw setpoint ($\pi$~rad/s). It also
extends the signature with a feed-forward acceleration term
\texttt{target\_acc}, which the cascaded PID adds to the position-loop
output before computing the thrust direction. Empirically
\texttt{DSLPIDControl} is preferred for aggressive trajectory tracking
(e.g., the waypoint square in \texttt{pid\_velocity}), while
\texttt{PIDControl} is preferred for hover-grade tasks.

When \texttt{act\_type=ActionType.PID} (or \texttt{VEL}, or
\texttt{ATTITUDE}) the environment runs the chosen controller
internally each \texttt{env.step}, so the agent can supply waypoints,
velocities, or attitude setpoints directly; this is identical in spirit
to the ``ctrl wrapper'' of \emph{gym-pybullet-drones} but is implemented
inline in \texttt{BaseAviary.\_preprocessAction}.

\subsubsection{SE(3) geometric controller for aggressive trajectories}
\begin{figure}[t]
  \centering
  \includegraphics[width=\columnwidth]{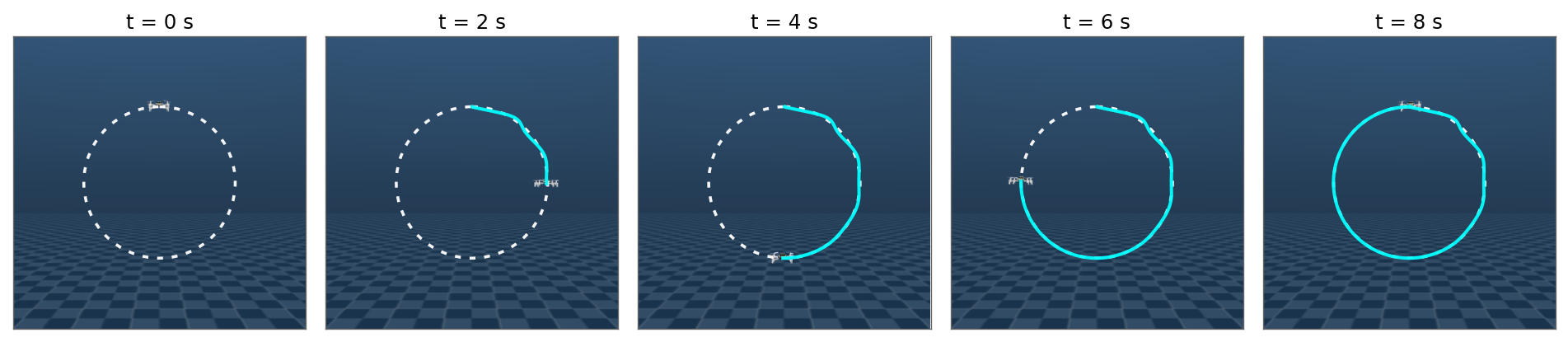}
  \caption{Vertical circle ($R=0.3$~m, $T=8$~s) in the $YZ$ plane
  executed by an SE(3) geometric controller~\cite{lee2010geometric}
  driving the real MuJoCo dynamics, rendered with the dedicated
  \texttt{front} camera. The dotted white ring is the planned
  trajectory, the cyan curve is the realised path; measured tracking
  error: mean $0.7$~cm, peak $3.5$~cm. The same camera mode is used by
  \texttt{RaceAviary} to visualise gate-racing trials.}
  \label{fig:vertical-circle}
\end{figure}

For high-bandwidth trajectory tracking we additionally provide a
classical SE(3) geometric controller~\cite{lee2010geometric} as a
drop-in replacement for the position PID. Figure~\ref{fig:vertical-circle}
shows a vertical circle of radius $0.3$~m and period $8$~s in the $YZ$
plane executed by this controller on the real MuJoCo dynamics; the
measured tracking error is $0.7$~cm on average ($\approx 2\%$ of $R$),
and peaks at $3.5$~cm ($\approx 12\%$ of $R$) at the apex. The default
\texttt{PIDControl} cannot match this on the same loop because a pure
position PID cannot supply the $v^2/R$ centripetal acceleration without
acceleration feed-forward.

\subsection{Examples}
\label{subsec:exp-examples}

All examples below default to $f_{\text{sim}} = 240$~Hz and
$f_{\text{ctrl}} = 48$~Hz unless noted.

\subsubsection{PID demos}
\texttt{examples/pid.py} contains three self-contained demos:
\texttt{pid\_hover} (single-drone hover with \texttt{Logger} output to
CSV, used to generate Fig.~\ref{fig:hover-track}), \texttt{pid\_velocity}
(square-waypoint chasing with \texttt{DSLPIDControl}), and
\texttt{multi\_drone\_pid} (three drones at staggered altitudes under
\texttt{Physics.MJC\_GND\_DRAG\_DW}). The matching downwash demo of
Fig.~\ref{fig:downwash} is reproduced by \texttt{examples/downwash.py}.

\subsubsection{Single-agent PPO}
\texttt{examples/learn.py} trains a
Stable-Baselines3~\cite{raffin2021stable} PPO agent on
\texttt{HoverAviary} with a 4-way \texttt{SubprocVecEnv} and an
\texttt{EvalCallback}; \texttt{examples/train\_rl.py} is its minimal
scalar version. \\\texttt{examples/play.py} loads a trained policy and
renders \texttt{--episodes} rollouts to PNGs or an animated GIF:

{\footnotesize
\begin{verbatim}
python examples/learn.py                    # single agent
python examples/learn.py --multiagent true  # N=2
python examples/play.py \
  --model_path results/rl_hover/best_model.zip \
  --env_type   hover --episodes 3
\end{verbatim}
}

\texttt{examples/train\_all\_envs.py} iterates over
\texttt{HoverAviary}, \texttt{VelocityAviary},
\texttt{FlyThroughAviary}, \texttt{RaceAviary},
\texttt{MultiHoverAviary}, and \texttt{FormationAviary} and is useful as
a smoke test of every env on a new machine.

\subsubsection{Logging and tests}
The \texttt{Logger(num\_drones, logging\_freq, output\_folder)} utility
in \texttt{multi\_drone\_mujoco.utils.logger} accumulates per-drone
$(t, \mathbf{p}, \mathbf{q}, \bm{\eta}, \mathbf{v}, \bm{\omega},
\mathbf{a})$ tuples and writes one CSV per drone via
\texttt{save\_to\_csv}; a companion \texttt{plot} method produces the
customary four-panel matplotlib time series. The \texttt{tests/}
directory ships pytest suites covering all seven envs
(\texttt{test\_envs.py}), the PID and DSL-PID controllers
(\texttt{test\_control.py}), the PettingZoo contract
(\texttt{test\_multi\_agent.py}, auto-skipped if PettingZoo is
unavailable), and the wind / domain-randomization / obstacle /
curriculum wrappers (\texttt{test\_features.py}). The suite runs on
any machine that can \texttt{import mujoco}; no GPU is required.
\section{Conclusions and Future Work}
\label{sec:conclusions}

We presented \textbf{MuJoCo-drones-gym}, an open-source, Gymnasium- and
PettingZoo-compatible multi-quadrotor environment built on top of the
MuJoCo physics engine. MuJoCo-drones-gym preserves the API and the
research-oriented spirit of
\emph{gym-pybullet-drones}~\cite{pybullet-drones},
while taking advantage of MuJoCo's contact handling, integrator
quality, and modern Python tooling. It ships with seven task templates,
five interchangeable action interfaces, three observation types, six
physics modes (with toggleable ground effect, drag, and downwash),
three drone models, composable wrappers for wind, obstacles,
curriculum, and domain randomization, and a GPU-vectorized MJX back end
for batched JAX-native training. Table~\ref{tab:mj-vs-pyb} summarizes
the head-to-head with gym-pybullet-drones; by design, the aerodynamic
models, drone parameters, and PID architecture mirror its predecessor
so that controllers and policies port across the two with minimal
effort.

\begin{table}[t]
\centering
\caption{Feature comparison between \emph{gym-pybullet-drones} and
\emph{MuJoCo-Drones-Gym}.}
\label{tab:mj-vs-pyb}
\renewcommand{\arraystretch}{1.25}
\setlength{\tabcolsep}{6pt}
\small
\begin{tabular}{@{}l p{0.30\textwidth} p{0.48\textwidth}@{}}
\toprule
\textbf{Feature} & \textbf{gym-pybullet-drones} & \textbf{MuJoCo-Drones-Gym}\\
\midrule
Engine        & PyBullet                    & MuJoCo $\geq 3.0$ \\
Renderer      & PyBullet GUI / TinyRenderer & \texttt{mujoco.viewer} + \texttt{mujoco.Renderer} \\
Drone models  & \texttt{CF2X}, \texttt{CF2P} & $+$ \texttt{RACE} (250\,g) \\
Task envs     & \texttt{Hover}, \texttt{MultiHover}
              & $+$ \texttt{Velocity}, \texttt{FlyThrough},
                \texttt{Formation}, \texttt{Race}, \texttt{MultiAgent} \\
Action types  & RPM, normalized            & $+$ \texttt{VEL}, \texttt{PID}, \texttt{ATTITUDE} \\
Aerodynamics  & drag, ground, downwash     & identical formulae, toggleable \\
Wind          & --                         & \texttt{CONSTANT}\,/\,\texttt{GUST}\,/\,\texttt{DRYDEN}\,/\,\texttt{SINUSOIDAL}\,/\,\texttt{COMBINED} \\
Obstacles     & --                         & procedural \texttt{FOREST}\,/\,\texttt{URBAN}\,/\,\texttt{INDOOR}\,/\,\texttt{RANDOM}\,/\,\texttt{GATES}\,/\,\texttt{CUSTOM} \\
Curriculum    & --                         & metric-driven \texttt{CurriculumWrapper} \\
MARL API      & custom subclass            & PettingZoo \texttt{ParallelEnv} \\
GPU vector.   & --                         & MuJoCo MJX via JAX \\
Assets        & submodule                  & vendored under \texttt{assets/cf2/}, env-overridable \\
\bottomrule
\end{tabular}
\end{table}

We see several promising avenues for future work. First, the existing
GPU-vectorized \\\texttt{MJXVectorAviary} back end currently supports
rigid-body MuJoCo physics only; extending it with XLA-compilable
versions of the ground-effect, drag, and downwash models, as well as
with multi-drone and vision-based observations, would close the
feature gap with the CPU \texttt{BaseAviary} and enable end-to-end
PureJaxRL- and Brax-style training pipelines on the full task suite.
Second, the existing Dryden wind, domain-randomization, and
curriculum-learning wrappers form a natural starting point for
sim-to-real transfer experiments on real Crazyflie~2.x hardware,
ideally accompanied by a flight-log dataset to identify residual model
error. Third, the racing and formation environments can be extended
with richer perception (stereo depth, event cameras) and richer reward
shaping, turning the package into a benchmark for vision-based
quadrotor RL. Finally, we plan to release a set of reference policies
and pretrained checkpoints, together with reproducible Colab
notebooks, to lower the barrier of entry for researchers and students
entering the field.

MuJoCo-drones-gym is released under the same license as the upstream
repository at
\url{https://github.com/tau-intelligence/MuJoCo-drones-gym}; contributions,
issues and pull requests are welcome.

\section*{Acknowledgements}
This work builds on the design and API conventions established by
\emph{gym-pybullet-drones}~\cite{pybullet-drones}. We gratefully
acknowledge the developers of \emph{MuJoCo}~\cite{todorov2012mujoco} and the
\emph{MuJoCo Menagerie}~\cite{menagerie2022} for the Bitcraze Crazyflie~2.x
MJCF model, as well as Bitcraze for the open hardware and firmware that have
made the Crazyflie a community standard for quadcopter research.

\label{section: References}
\bibliographystyle{IEEEtran}
\bibliography{references.bib}

\end{document}